\documentclass[review]{elsarticle}

\usepackage{lineno,hyperref}
\date{}
\usepackage{amsmath,amsfonts}
\usepackage[resetlabels,labeled]{multibib}
\usepackage{graphicx}
\usepackage{float}
\usepackage{epsfig}  
\usepackage{multirow}
\usepackage[colorinlistoftodos]{todonotes}
\usepackage{subcaption}
\usepackage{times}
\usepackage{url}
\usepackage{latexsym}
\usepackage{times}
\usepackage{comment}
\usepackage{multirow}
\usepackage{array}
\usepackage[utf8]{inputenc}
\usepackage{color}
\usepackage{amsmath}
\usepackage{cleveref}
\crefname{section}{§}{§§}
\Crefname{section}{§}{§§}

\DeclareMathOperator*{\argmax}{argmax}

\usepackage{parallel,enumitem}
\usepackage{subcaption}

\modulolinenumbers[5]

\journal{Journal of \LaTeX\ Templates}

\begin{document}

\begin{frontmatter}

\title{Unified Neural Architecture for Drug, Disease and Clinical Entity Recognition}
%\tnotetext[mytitlenote] {Fully documented templates are available in the elsarticle package on \href{http://www.ctan.org/tex-archive/macros/latex/contrib/elsarticle}{CTAN}.}

%% Group authors per affiliation:
\author{Sunil Kumar Sahu, Ashish Anand}
\address{Department of Computer Science and Engineering \\ Indian Institute of Technology Guwahati, India \\ {\tt \{sunil.sahu, anand.ashish\}@iitg.ernet.in}}
%\fntext[myfootnote]{Since 1880.}

%% or include affiliations in footnotes:
%\author[mymainaddress,mysecondaryaddress]{Elsevier Inc}
%\ead[url]{www.elsevier.com}

%\author[mysecondaryaddress]{Global Customer Service\corref{mycorrespondingauthor}}
%\cortext[mycorrespondingauthor]{Corresponding author}
%\ead{support@elsevier.com}

%\address[mymainaddress]{1600 John F Kennedy Boulevard, Philadelphia}
%\address[mysecondaryaddress]{360 Park Avenue South, New York}

\begin{abstract}
Most existing methods for biomedical entity recognition task rely on explicit feature engineering where many features either are specific to a particular task or depends on output of other existing NLP tools. Neural architectures have been shown across various domains that efforts for explicit feature design can be reduced. In this work we propose an unified framework using bi-directional long short term memory network (BLSTM) for named entity recognition (NER) tasks in biomedical and clinical domains. Three important characteristics of the framework are as follows - (1) model learns contextual as well as morphological features using two different BLSTM in hierarchy, (2) model uses first order linear conditional random field (CRF) in its output layer in cascade of BLSTM to infer label or tag sequence, (3) model does not use any domain specific features or dictionary, i.e., in another words, same set of features are used in the three NER tasks, namely, disease name recognition ({\it Disease NER}), drug name recognition ({\it Drug NER}) and clinical entity recognition ({\it Clinical NER}). We compare performance of the proposed model with existing state-of-the-art models on the standard benchmark datasets of the three tasks. We show empirically that the proposed framework outperforms all existing models. Further our analysis of CRF layer and word-embedding obtained using character based embedding show their importance.
%  Through the experiments done on disease name recognition ({\it Disease NER}), drug name recognition ({\it Drug NER}) and clinical entity recognition ({\it Clinical NER}) tasks we demonstrate the importance of proposed method.
\end{abstract}

\begin{keyword}
\texttt{Drug Name Recognition, Disease Name Recognition, Clinical Entity Recognition, Recurrent Neural Network, LSTM Network} 
\end{keyword}

\end{frontmatter}

\section{Introduction}
Biomedical and clinical named entity recognition (NER) in text is one of the important step in several biomedical and clinical information extraction tasks \cite{Rosario04,segura2015exploring,uzuner2010}. State-of-art methods formulated NER task as a sequence labeling problem where each word is labeled with a tag and based on tag sequence entities of interest get identified. It has been observed that named entity recognition in biomedical and clinical domain is difficult \cite{leaman09,uzuner10a} compared to the generic domain. There are several reasons behind this, including use of non standard abbreviations or acronyms, multiple variations of same entities etc. Further clinical notes are more noisy, grammatically error prone and contain less context due to shorter and incomplete sentences \cite{uzuner2010}. Most widely used models such as CRF, maximum entropy Markov model (MEMM) or support vector machine (SVM), use manually designed rules to obtain morphological, syntactic, semantic and contextual information of a word or of a piece of text surrounding a word, and use them as features for identifying correct label \cite{Lafferty:2001,MahbubChowdhury10,jiang2011study,rocktaschel2013wbi,bjorne2013uturku}. It has been observed that performance of such models are limited with the choice of explicitly designed features which are generally specific to task and its corresponding domain. For example, Chawdhury and Lavelli \cite{MahbubChowdhury10} explained several reasons why features designed for biological entities such as protein or gene are not equally important for disease name recognition.

Deep learning based models have been used to reduce manual efforts for explicit feature design in \cite{collobert11a}. Here distributional features were used in place of manually designed features and multilayer neural network were used in place of linear model to overcome the needs of task specific meticulous feature engineering. Although proposed methods outperformed several generic domain sequence tagging tasks but it fails to overcome state-of-art in biomedical domain \cite{LinYao2015}. There are two plausible reasons behind that, first, it learned features only from a word level embedding and second, it took into account only a fixed length context of the word. It has been observed that word level embeddings preserve syntactic and semantic properties of word, but may fail to preserve morphological information which can also play important role in biomedical entity recognition \cite{dos2014,lample2016neural,MahbubChowdhury10,LeamanG08}. For instance, drug names {\it Cefaclor, Cefdinir, Cefixime, Cefprozil, Cephalexin} have common prefix and {\it Doxycycline, Minocycline, Tetracycline} have common suffix. Further, window based neural architecture can only consider contexts falling within the user decided window size and will fail to pick important clues lying outside the window.
% * <anand.ashish@iitg.ernet.in> 2017-07-07T03:21:27.834Z:
%1. should give reference to support the argument "word-embeding fail to preserve morphological information.
% 2. But collobert paper also considers sentence level.
% ^.

This work aims to overcome the above mentioned two issues. The first one to obtain both morphologically  as well as syntactic and semantically rich embedding, two BLSTMs are used in hierarchy. First BLSTM works on each character of words and obtain morphologically rich word embedding. Second BLSTM works at word level of a sentence to learn contextually reach feature vectors. The second one to make sure all context lying anywhere in the sentence should be utilized, we consider entire sentence as input and use first-order linear chain CRF in the final prediction layer. The CRF layer accommodates dependency information about tags. We evaluate the proposed model on three standard biomedical entity recognition tasks namely {\it Disease NER}, {\it Drug NER} and {\it Clinical NER}. To the best of our knowledge this is the first work which explores single model using character based word embedding in conjunction with word embedding for drug and clinical entity recognition tasks. We compare the proposed model with the existing state-of-the-art models for each task and show that it outperforms them. Further analysis of the model indicates the importance of using character based word embedding along with word embedding and CRF layer in the final output layer.

\section{Method}
\label{ner_method}

\subsection{Bidirectional Long Short Term Memory}
\label{sec:rnn}
Recurrent neural network (RNN) is a variant of neural networks which utilizes sequential information and maintains history through its recurrent connection \cite{Graves:2009,Graves13}. RNN can be used for a sequence of any length, however in practice it fails to maintain long term dependency due to vanishing and exploding gradient problems~\cite{bengio2013,bengio2013advances}. Long short term memory (LSTM) network \cite{Hochreiter97} is a variant of RNN which takes care of the issues associated with vanilla RNN by using three gates (input, output and forget) and a memory cell. 

We formally describe the basic equations pertaining to LSTM model. Let $h^{(t-1)}$ and $c^{(t-1)}$ be hidden and cell states of LSTM respectively at time $t-1$, then computation of current hidden state at time $t$ can be given as:
\begin{align*} 
&i^{(t)} = \sigma ( U^{(i)} x^{(t)} +  W^{(i)} h^{(t-1)} + b^i)\\
&f^{(t)} = \sigma (U^{(f)} x^{(t)} + W^{(f)} h^{(t-1)} + b^f)\\
&o^{(t)} = \sigma (U^{(o)} x^{(t)}  + W^{(o)} h^{(t-1)} + b^o)\\
&g^{(t)} = tanh(U_l^{(g)} x^{(t)} +  W^{(g)} h^{(t-1)} + b^{g}) \\
&c^{(t)} = c^{(t-1)} * f^{(t)} + g^{(t)} * i^{(t)} \\
&h^{(t)} = tanh(c^{(t)}) * o^{(t)},
\end{align*} 
where $\sigma$ is sigmoid activation function, $*$ is an element wise product, $x^{(t)} \in \mathbb{R}^d$ is the input vector at time $t$, $U^{(i)}$, $U^{(f)}$, $U^{(o)}$, $U^{(g)} \in \mathbb{R}^{N \times d}$, $W^{(i)}$, $W^{(o)}$, $W^{(f)}$, $W^{(g)} \in \mathbb{R}^{N \times N}$, $b^i$, $b^f$, $b^o$, $b^g \in \mathbb{R}^{N}$, $h^{(0)}$, $c^{(0)} \in \mathbb{R}^N$ are learning parameters for LSTM. Here $d$ is dimension of input feature vector, $N$ is hidden layer size and $h^{(t)}$ is output of LSTM at time step $t$. 

It has become common practice to use LSTM in both forward and backward directions to capture both past and future contexts respectively. First LSTM computes its hidden states in forward direction of input sequence and second does it in backward direction. This way of using two LSTMs is referred to as bidirectional LSTM or simply BLSTM. We have also used bi-directional LSTM in our model. Final output of BLSTM at time $t$ is given as:
\begin{equation}
 h^{(t)} = \overrightarrow{h^{(t)}} \oplus \overleftarrow{h^{(t)}}
\end{equation}
Where $\oplus$ is concatenation operation and $\overrightarrow{h^{(t)}}$ and $\overleftarrow{h^{(t)}}$ are hidden states of forward and backward LSTM at time $t$.

\begin{figure}[!tb] 
\begin{center}
\includegraphics[width=0.9\textwidth]{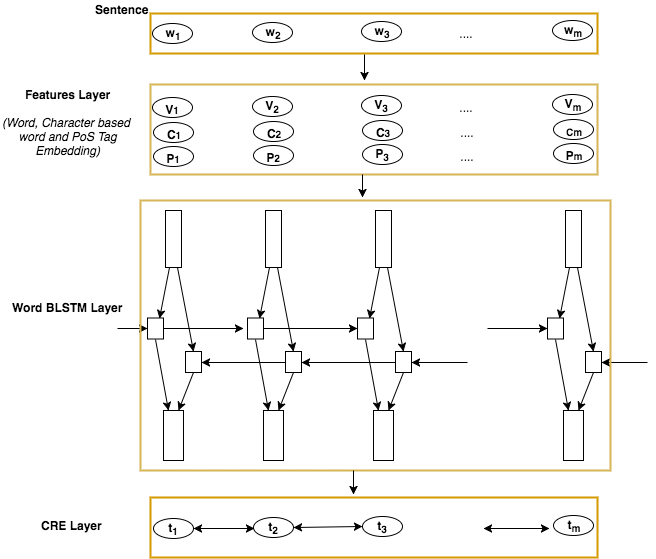}
\caption{Bidirectional recurrent neural network based model for biomedical entity recognition. Here $w_1 w_2 ... w_m$ is the word sequence of the  sentence and $t_1$ $t_2$ ... $t_m$ is its computed label sequence and $m$ represents length of the sentence.}
\label{fig:ner_model}
\end{center}
\end{figure}

\subsection{Model Architecture}
\label{sec:ner_model}
Similar to any named entity recognition task, we formulate the biomedical entity recognition task as a token level sequence tagging problem. We use BIO tagging scheme in our experiments \cite{settles2004}. Architecture of the proposed model is presented in Figure \ref{fig:ner_model}. Our model takes whole sentence as input and compute a label sequence as output. First layer of the model learns local feature vectors for each word in the sentence. We use concatenation of word embedding, PoS tag embedding and character based word embedding as a local feature for every word. Character based word embedding is learned through applying a BLSTM on the character vectors of a word. We call this layer as {\it Char BLSTM} (~\cref{sec:crnn} ). Subsequent layer, called \textit{Word BLSTM} (~\cref{sec:crf}), incorporates contextual information on it through a separate BLSTM network. Finally we use a CRF to encode correct label sequence on the output of {\it Word BLSTM} (~\cref{sec:crf} ). Now onwards, the proposed framework will be referred to as \textit{CWBLSTM}. Entire network parameters are trained in end-to-end manner through cross entropy loss function. We next describe each part of the model in detail.

\subsection{Features Layer}
\label{sec:feat}
Word embedding or distributed word representation is a compact vector represent of a word which preserve lexico-semantic properties \cite{Bengio03}. 
It is a common practice to initialize word embedding with a pre-trained vector representation of words.  
Apart from word embedding in this work PoS tag and character based word embedding are used as features. The output of feature layer is a sequence of vectors say $x_1, \cdots x_m$ for the sentence of length $m$. Here $x_i \in \mathbb{R}^d$ is the concatenation of word embedding, PoS tag embedding and character based word embedding. We next explain how character based word embedding is learned.

\subsubsection{Char BLSTM}
\label{sec:crnn}
Word embedding is crucial component for all deep learning based NLP task. Capability to preserve lexico-semantic properties in vector representation of a word made it a powerful resource for NLP \cite{collobert11a,Turian10}. In biomedical and clinical entity recognition tasks apart from semantic information, morphological structure such as prefix, suffix or some standard patterns of words also give important clues \cite{MahbubChowdhury10,leaman09}. The motivation behind using character based word embedding is to incorporate morphological information of words in feature vectors.

To learn character based embeddings, we maintained a vector for every characters in a embedding matrix \cite{dos2014,lample2016neural}. These vectors are initialized with random values in the beginning. To illustrate, suppose {\it cancer} is a word for which we want to learn an embedding (represented in figure \ref{fig:charRNN}), we use a BLSTM on the vector of each characters of {\it cancer}. As mentioned earlier forward LSTM maintained information about past in computation of current hidden state and backward LSTM obtained futures contexts, therefore after reading entire sequence, last hidden states of both RNN must have knowledge of whole word with respect to their directions. The final embedding of a word would be: 
\begin{figure}[!tb] 
\begin{center}
\includegraphics[width=0.6\textwidth]{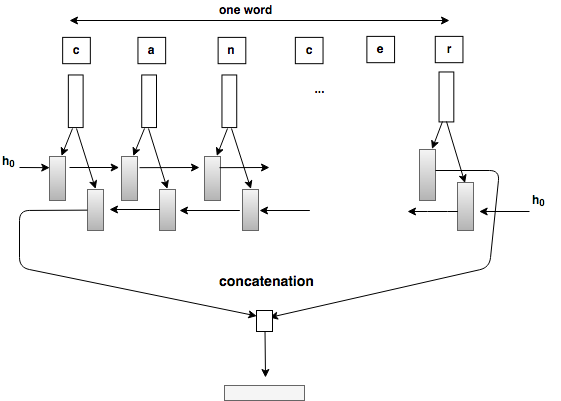}
\caption{Learning character based word embedding}
\label{fig:charRNN}
\end{center}
\end{figure}
\begin{equation}
v_{cw} = \overrightarrow{h^{(m)}} \oplus \overleftarrow{h^{(m)}}
\end{equation}
Where $\overrightarrow{h^{(m)}}$ and $\overleftarrow{h^{(m)}}$ is the last hidden states of forward and backward LSTMs respectively.

\subsection{Word BLSTM Layer}
\label{sec:global}
The output of feature layer is a sequence of vectors for each word of the sentence. These vectors have local or individual information about the words. Although local information plays important role in identifying entities, but a word can have different meaning in different contexts. Earlier works in \cite{collobert11a,LeamanG08,MahbubChowdhury10,LinYao2015} use a fixed length window to incorporate contextual information. However important clues can lie anywhere in the whole sentence. This limit the learned vectors to obtain knowledge about complete sentence. To overcome this, we use a separate BLSTM network which takes local feature vectors as input and outputs a vector for every word based on both contexts and current feature vectors. 

\subsection{CRF Layer}
\label{sec:crf}
The output of {\it Word BLSTM} layer is again a sequence of vectors which have contextual as well as local information. One simple way to decode the feature vector of a word into its corresponding tag is to use word level log likelihood (WLL) \cite{collobert11a}. Similar to {\it MEMM's} it will map the feature vector of a word to a score vector of each tag by a linear transformation and every word will get its label based on its scores and independent of labels of other words. One limitation of this way of decoding is it does not take into account dependency among tags. For instance in {\it BIO tagging} scheme a word can only be tagged with {\it I-Entity} (standing for Intermediate-Entity) only after a {\it B-Entity} (standing for Beginning-Entity). We use CRF \cite{Lafferty:2001} on the feature vectors to include dependency information in decoding and decode whole sentence together with its tag sequence.

CRF maintained two parameters for decoding, $W_{u} \in R^{k\times h}$ linear mapping parameter and $T \in R^{h\times h}$ pairwise transition score matrix. Here $k$ is the size of feature vector, $h$ is the number of labels present in task and $T_{i,j}$ implies pair wise transition score for moving from label $i$ to label $j$. Let $[v]_{1}^{|s|}$ be a sequence of feature vectors for a sentence $[w]_{1}^{|s|}$ and suppose $[z]_{1}^{|s|}$ is the unary potential scores obtained after applying linear transformation on feature vectors (here $z_i \in R^h$) then CRF decodes this with tag sequence using:
\begin{equation}
P( [y]_{1}^{|s|} | [w]_{1}^{|s|} ) = \argmax_{t\in Q^{|s|}} \frac{ \exp \Psi ( [z]_{1}^{|s|} , [t]_{1}^{|s|})} { \sum_{t^{\psi} \in Q^{|s|}} \exp \Psi ( [z]_{1}^{|s|} , [t^{\psi}]_{1}^{|s|} ) } 
\end{equation} 
where 
\begin{equation}
\Psi( [z]_{1}^{|s|} , [t]_{1}^{|s|}) = \sum_{1 \le i \le |s|} (T_{t_{i-1},t_{i}} + z_{t_i})
\end{equation} 
Here $Q^{|s|}$ is a set contain all possible tag sequence of length $|s|$, $t_j$ is tag for the $j^{th}$ word. Highest probable tag sequence is estimated using Viterbi algorithm \cite{rabiner1989,collobert11a}. 

\subsection{Training and Implementation}
We use cross entropy loss function to train the model. Adam's technique \cite{adam2014} is used for updating entire neural network and embedding parameters of our model. We use mini batch size of $50$ in training for all tasks. Entire implementation is done in python language using {\it Tensorflow}\footnote{https://www.tensorflow.org} package. In all our experiments, we use pre-trained word embedding of length $100$, which was trained on PubMed corpus using GloVe \cite{pennington14,muneeb15}, PoS tag embedding vector of length $10$ and character based word embedding of length $20$.  We used $l_2$ regularization with $0.001$ as corresponding parameter value. These hyperparameters are obtained using validation set of {\it Disease NER} task. The corresponding training, validation and test sets for \textit{Disease NER} task is available as separate files with NCBI disease corpus. For the other two tasks, we used the same set of hyperparameters as obtained on \textit{Disease NER}.
% * <anand.ashish@iitg.ernet.in> 2017-07-07T05:01:33.434Z:
% 
% whether any tuning of the hyper-parameters done?
% 
% Sunil: All the hyperparameters are obtained based on validation set of disease NER task which was a separate file shared by NCBI 
% ^.
\section{The Benchmark Tasks}
\label{sec:dataset}
In this section, we briefly describe the three standard tasks on which we examine the CWBLSTM model. Statistics of corresponding benchmark datasets is given in Table \ref{tab:ner_stats}.

\subsection{Disease NER} 
Identifying disease named entity in text is crucial for disease related knowledge extraction~\cite{bundschus2008,agarwal2008}. Furthermore, It has been observed that disease is one of the most widely searched entities by users on PubMed \cite{Dogan12}. We use {\it NCBI disease corpus}\footnote{https://www.ncbi.nlm.nih.gov/CBBresearch/Dogan/DISEASE/} to investigate performance of the model on {\it Disease NER} task. This dataset was annotated by a team of $12$ annotators (2 persons per annotation) on 793 PubMed abstracts \cite{Dogan12,Dogan14}. 
% This dataset is an extension of the AZDC dataset \cite{leaman09}. As part of extension NCBI disease corpus contained all sentences from considered PubMed citations however, AZDC dataset included only sentences where disease name appeared. 

\subsection{Drug NER}
Identifying drug name or pharmacological substance are important first step for drug drug interaction extraction and for other drug related knowledge extraction tasks. Keeping this in mind a challenge for recognition and classification of pharmacological substances in text was organized as part of SemEval 2013. We use SemEval-2013 task 9.1 \cite{segura2013} dataset for this task. The dataset shared in this challenge were annotated from two sources {\it DrugBank}\footnote{https://www.drugbank.ca/} documents and {\it MedLine}\footnote{https://www.nlm.nih.gov/bsd/pmresources.html} abstracts.  This dataset has four kind of drugs as entities, namely {\it drug}, {\it brand}, {\it group} and {\it drug\_n}. Here {\it drug} represent generic drug name, {brand} is brand name of a drug, {\it group} is family name of drugs and {\it drug\_n} is active substance not approved for human use \cite{segura2011chal}. In this case while processing the dataset, $79$ entities ($56$ {\it drug}, $18$ $group$ and $5$ $brand$) from training set and $5$ entities ($4$ $drug$ and $1$ $group$) from test set were missed. Missed entities of test set are treated as false negative in our evaluation scheme.

\begin{table}[t]
\centering
\scalebox{0.99}{
\begin{tabular}{|c|c|c|c|c|}
\hline
\textit{\textbf{Dataset}} & \textit{\textbf{Corpus}} & \textit{\textbf{Train set}} & \textit{\textbf{Test set}} \\ \hline
\multirow{2}{*}{Disease NER} 
& sentences  	& 5661 & 961 \\ \cline{2-4}
& disease 		& 5148 & 961 \\ 
\hline
\hline
\multirow{5}{*}{Drug NER} 
 & sentences  	& 6976 & 665  \\ \cline{2-4}
 & drug  		& 9369 & 347 \\  \cline{2-4}
 & brand  		& 1432 & 59  \\  \cline{2-4}
 & group  		& 3381 & 154  \\ \cline{2-4}
 & drug\_n  	& 504  & 120  \\ 
\hline
\hline
\multirow{4}{*}{Clinical NER} 
 & sentences  	& 8453 &  14529 \\ \cline{2-4}
 & problem  	& 7072 &  12592 \\ \cline{2-4}
 & treatment  	& 2841 &  9344  \\ \cline{2-4}
 & test  		& 4606 &  9225  \\ \cline{2-4}
 \hline
\end{tabular}
}
\caption{Statistics of benchmark datasets for the three tasks used in the study.}
\label{tab:ner_stats}
\end{table}

\subsection{Clinical NER}
For clinical entity recognition we used publicly available (under license) i2b2/VA\footnote{https://www.i2b2.org/NLP/Relations/Main.php} challenge dataset \cite{uzuner2010,uzuner10a}. This dataset is a collection of discharge summaries obtained from Partners Healthcare, Beth Israel Deaconess Medical Center, and the University of Pittsburgh Medical Center. The dataset was annotated for three kinds of entities namely {\it problem}, {\it treatment} and {\it test}. Here \textit{problems} indicate phrases that contain observations made by patients or clinicians about the patient’s body or mind that are thought to be abnormal or caused by a disease. \textit{Treatments} are phrases that describe procedures, interventions, and substances given to a patient in an effort to resolve a medical problem. \textit{Tests} are procedures, panels, and measures that are done to a patient or a body fluid or sample in order to discover, rule out, or find more information about a medical problem. 

The downloaded dataset for this task was only partially available (only discharge summaries from Partners Healthcare and Beth Israel Deaconess Medical Center) compared to the full dataset originally used in the challenge. We performed our experiments on currently available partial dataset. The dataset is available in pre-processed form, where sentence and word segmentations were alredy done. We removed patient's information from each discharge summary before training and testing, because that never contains entities of interest.
 
\section{Results and Discussion}

\subsection{Experiment Design}
We perform separate experiments for each task. We use {\it train set} for learning optimal parameters of the model for each dataset and evaluation is performed on {\it test set}. Performance of each trained model is evaluated based on strict matching sense, where exact boundaries as well as class need to be correctly identified for consideration of true positive. For strict matching evaluation scheme, we use CoNLL 2004\footnote{http://www.cnts.ua.ac.be/conll2002/ner/bin/conlleval.txt} evaluation script to calculate precision, recall and F1 score in each task.

\subsection{Baseline Methods}
We use following methods as a common baseline for comparison with the proposed models in all of the considered tasks. The selected baseline methods are implemented by us: 

%\begin{itemize}
%\item 
{\bf SENNA:} SENNA uses window based neural network on embedding of a word with its context to learn global feature \cite{collobert11a}. To make inference it also uses CRF on the output of window based neural network. We set the window size $5$ based on hyperparameter tuning using validation set ($20\%$ of training set), and rest all the hyperparameters are set similar to our model.

%\item 
{\bf CharWNN:} This model~\cite{dos2014} is similar to SENNA but uses word as well as character based embedding in the chosen context window \cite{dos2015}. Here character based embeddings are learned through convolution neural network with max pooling scheme.

%\item 
{\bf CharCNN:} This method~\cite{sunil16b} is similar to the proposed model \textit{CWBLSTM} but instead of using BLSTM, it uses convolution neural network for learning character based embedding.
%\end{itemize}

\subsection{Comparison with Baseline} 
\begin{table}[t]
\centering
\scalebox{0.99}{
\begin{tabular}{|c|l|c|c|c|c|}
\hline
{\textbf{Tasks}} & {\textbf{Models}}& {\textbf{Accuracy}} & {\textbf{Precision}} & {\textbf{Recall}} & {\textbf{F1 Score}}  \\ \hline
\hline
\multirow{4}{*}{\textbf{Disease NER}}
& {\it SENNA}					& 97.26 & 77.93 & 76.80 & 77.36 \\ \cline{2-6}
& {\it CharWNN} 				& 97.24 & 78.34 & 78.67 & 78.50 \\ \cline{2-6}
& {\it CharCNN} 				& 97.61 & 84.26 & 78.56 & 81.31 \\ \cline{2-6}
& {\it CWBLSTM}				& {\bf 97.77} & {\bf 84.42} & {\bf 82.31} & {\bf 83.35} \\ \cline{2-6}
%& {\it Bi-RNN+CharCNN+CharRNN} 	& 97.75 & 81.53 & 80.85 & 81.19  \\ 
\hline 
\hline 
\multirow{4}{*}{\textbf{Drug NER}}
& {\it SENNA}					& 96.71 & 66.93 & 62.70 & 64.75 \\ \cline{2-6}
& {\it CharWNN} 				& 97.07 & 69.16 & 69.16 & 69.16 \\ \cline{2-6}
& {\it CharCNN} 				& 97.09 & 70.34 & 72.10 & 71.21 \\ \cline{2-6}
& {\it CWBLSTM}				& {\bf 97.46} & {\bf 72.57} & {\bf 74.60} & {\bf 73.57} \\ \cline{2-6}
%& {\it Bi-RNN+CharCNN+CharRNN} 	& {\bf 97.40} & {\bf 73.78} & {\bf 73.13} & {\bf 73.45}  \\ 
\hline 
\hline 
\multirow{4}{*}{\textbf{Clinical NER}}
& {\it SENNA}					& 91.56 & 80.30 & 78.85 & 79.56 \\ \cline{2-6}
& {\it CharWNN} 				& 91.42 & 79.96 & 78.12 & 79.03 \\ \cline{2-6}	
& {\it CharCNN} 				& 93.02 & 83.65 & 83.25 & 83.45  \\ \cline{2-6}
& {\it CWBLSTM}				& {\bf 93.19} & {\bf 84.17} & {\bf 83.20} & {\bf 83.68} \\ \cline{2-6}
%& {\it Bi-RNN+CharCNN+CharRNN}  & 93.13  & {\bf 84.74} & {\bf 83.07} & {\bf 83.90}  \\ \cline{2-6}
\hline
\end{tabular}
}
\caption{Performance comparison of the proposed model \textit{CWBLSTM} with baseline models on test set of different datasets. Here \textit{Accuracy} represents token level accuracy in tagging.}
\label{tab:comp_base}
\end{table}
Table \ref{tab:comp_base} presents comparison of \textit{CWBLSTM} with different baseline methods on disease, drug and clinical entity recognition tasks. We can observe that it outperforms all three baselines in each of the three tasks. In particular, when comparing with \textit{CharCNN}, differences are significant for \textit{Drug NER} and \textit{Disease NER} tasks but difference is insignificant for \textit{Clinical NER}. The proposed model improved the recall by $5\%$ to gain about $2.5\%$ of relative improvement in F1 score over the second best method \textit{CharCNN} for the \textit{Disease NER} task. For the \textit{Drug NER} task, relative improvement of more than $3\%$ is observed for all three measures, precision, recall and F1 score over the \textit{CharCNN} model. The relatively weaker performance on \textit{Clinical NER} task could be attributed to use of many non standard acronyms and abbreviations which makes it difficult for character based embedding models to learn appropriate representation. 
% This observation can also be obtained by observing performance difference of {\it SENNA} and {\it CharWNN} in this task. 

One can also observe that, even though {\it Drug NER} has sufficiently enough training dataset, all models gave relatively poor performance compared to the performance in other two tasks. One reason for the poor performance could be the nature of the dataset. As discussed {\it Drug NER} dataset constitutes texts from two sources, {\it DrugBank} and {\it MedLine}. Sentences from \textit{DrugBank} are shorter and are comprehensive as written by medical practitioners, whereas \textit{MedLine} sentences are from research articles which generally tend to be longer. Further the \textit{training set} constitutes $5675$ sentences from {\it DrugBank} and $1301$ from {\it MedLine}, whereas \textit{test set} this distribution is reversed, i.e. more sentences are from \textit{MedLine} ($520$ in comparison to $145$ sentences from \textit{DrugBank}). Smaller set of training instances from \textit{MedLine} sentences do not give sufficient examples to model to learn.

\begin{table}[tbp]
\centering
\scalebox{0.9}{
\begin{tabular}{|l|c|c|c|c|}
\hline
{\textbf{Model}} &{\bf Features} &{\textbf{Precision}} & {\textbf{Recall}} & {\textbf{F Score}} \\ \hline \hline
\textit{CWBLSTM} 	  & Word, PoS and Character Embedding & 84.42   & {\bf 82.31} & {\bf 83.35} \\ \hline
{\it BANNER\cite{Dogan12}}&	Orthographic, morphological, syntactic & 	-	& 	- 	 & 81.8 	\\ \hline
%{\it Bi-RNN+CharCNN \cite{sunil16b}} & & 76.98   &  75.80 & 76.30  \\ \hline
%{\it Bi-RNN+CharCNN+We\cite{sunil16b}}& &76.10   &  74.11 & 75.09  \\ \hline
{\it BLSTM+We\cite{sunil16b}}	& Word Embedding  &{\bf 84.87}& 74.11& 79.13	\\ \hline
\end{tabular}
}
\caption{Performance comparison of \textit{CWBLSTM} with other existing models on \textit{Disease NER} task}
\label{tab:sta_disease}
\end{table}

\subsection{Comparison with Other Methods}
In this section we compare our results with other existing methods present in literature. We do not compare results on \textit{Clinical NER} as the complete dataset (as was available in i2b2 challenge) is not available and results in literature are with respect to the complete dataset.

\subsubsection*{Disease NER}
Table \ref{tab:sta_disease} shows performance comparison of different existing methods with \textit{CWBLSTM} on NCBI disease corpus. \textit{CWBLSTM} improved the performance of BANNER by $1.89\%$ in terms of F1 Score. BANNER is a CRF based method which primarily uses orthographic, morphological and shallow syntactic features \cite{LeamanG08}. Many of these features are specially designed for biomedical entity recognition tasks. The proposed model also gave better performance than another BLSTM based model~\cite{sunil16b} by improving recall by around $12\%$. BLSTM model in \citep{sunil16b} used BLSTM network with word embedding only whereas the proposed model make use of extra features in terms of PoS as well as character based word embeddings.

\subsubsection*{Drug NER}
Table \ref{tab:sta_drug} reports performance comparison on {\it Drug NER} task with submitted results in SemEval-2013 Drug Named Recognition Challenge \cite{segura2013}. \textit{CWBLSTM} outperforms the best result obtained in the challenge (WBI-NER\cite{rocktaschel2013wbi}) by a margin of $1.8\%$. 
% One should note that here we presented our result by considering $5$ missed entities in false negative. 
{\it WBI-NER} is the extension of ChemSpot chemical NER\cite{chemspot2012} system which is a hybrid method for chemical entity recognition. ChemSpot primarily uses features from dictionary to make sequence classifier using CRF. Apart from that WBI-NER also used features obtained from different domain dependent ontologies. Performance of the proposed model is better than LASIGE \cite{grego2013lasige} as well as UTurku \cite{bjorne2013uturku} system's by a significant margin. LASIGE is also a CRF based method and UTurku uses Turku Event Extraction System (TEES), which is a kernel based model for entity and relation exaction tasks.
\begin{table}[tbp]
\centering
\scalebox{0.9}{
\begin{tabular}{|l|c|c|c|c|}
\hline
{\textbf{Model}} &{\bf Features} &{\textbf{Precision}} & {\textbf{Recall}} & {\textbf{F Score}} \\ \hline \hline
\textit{CWBLSTM} 	& Word,PoS and Character Embedding & \textbf{72.57} & \textbf{74.05} & \textbf{73.30}	\\ \hline
%\textit{\cite{liu2015feature}}			& 88.37 & 72.01 & 79.36 \\ \hline
\textit{WBI\cite{rocktaschel2013wbi}} & ChemSpot and Ontologies & 73.40 & 69.80 & 71.5	\\ \hline
%\textit{NLM\_LHC} 		 				& 73.20 & 67.90 & 70.4 	\\ \hline
\textit{LASIGE\cite{grego2013lasige}} &	Ontology and Morphological	& 69.60 & 62.10 & 65.6 	\\ \hline
\textit{UTurku\cite{bjorne2013uturku}}& Syntactic and Contextual	& 73.70 & 57.90 & 64.8	\\ \hline
%\textit{UEM\_UC3M\cite{sanchez2015uem}} & 51.70 & 54.20 & 52.9 	\\ \hline
\end{tabular}
}
\caption{Performance comparison \textit{CWBLSTM} with other existing models submitted in SemEval-2013 \textit{Drug NER} task}
\label{tab:sta_drug}
\end{table}

\subsection{Feature Ablation Study}
We analyze importance of each feature type by performing feature ablation. The corresponding results are presented in Table~\ref{tab:effect_feature}.
In this table first row present performance of the proposed model using all feature types in all three tasks and second, third and fourth rows shows performance when character based word embedding, PoS tag embeddings and pre-trained word embedding are removed from the model subsequently. Removal of pre-trained word embedding implies use of random vectors in place of pre-trained vectors.

Through the table we can observe that after removal of character based word embedding, $3.6\%$, $ 5.8\%$ and $1.1\%$ relative decrements in F1 Score on {\it Disease NER} and {\it Drug NER} and {\it Clinical NER} tasks are observed. This demonstrate the importance of character based embedding. As mentioned earlier character based word embedding helps our model in two ways, first, it gives morphologically rich vector representation and secondly, through character based word embedding we can get vector representation for  OoV  (out of vocabulary) words also. OoV words are $9.9\%$, $13.85\%$ and $20.13\%$ in {Drug NER}, {\it Disease NER} and {\it Clinical NER} dataset respectively (shown in table \ref{tab:oov}). As discussed earlier this decrements are less in {\it Clinical NER} because of presence of acronyms and abbreviations in high frequency which does not allow model to take advantage of character based word embedding.  Through third row we can also observe that using PoS tag embedding as feature is not so crucial in all three tasks. This is because distributed word embedding implicitly preserve that kind of information. 
\begin{table}[tbp]
\centering
\scalebox{0.8}{
\begin{tabular}{|c|c|c|c|}
\hline
{\textbf{Model}} & {\textbf{Disease NER}} & {\textbf{Drug NER}} & {\textbf{Clinical NER}} \\ \hline \hline
\textit{\textbf{CWBLSTM}}	& (84.42), (82.31), (83.35) & (72.57), (74.60), (73.57) & (84.17), (83.20), (83.68) \\ \hline
\textit{\textbf{- CE}}			& (80.86), (80.02), (80.44) & (64.29), (75.62), (69.50) & (83.76), (81.74), (82.74) \\ \hline
\textit{\textbf{- (CE+PE)}} 				& (82.72), (77.73), (80.15) & (65.96), (73.42), (69.49) & (83.31), (80.51), (81.89) \\ \hline
\textit{\textbf{- (CE+PE+WE)}} 				& (79.66), (73.78), (76.61) & (65.40), (55.80), (60.22) & (79.53), (78.28), (78.90)\\ \hline
\end{tabular}
}
\caption{Table explain contribution of each features on different datasets. In every block (X), (Y), (Z) implies precision, recall and F score respectively. Here in row $4$ model uses random vector in place of pre-trained word vectors for word embedding.}
\label{tab:effect_feature}
\end{table}
\begin{table}[tbp]
\centering
\scalebox{0.9}{
\begin{tabular}{|c|c|c|c|}
\hline
{\bf Dataset} 					& {\bf Unique Words} & {\bf OoV } & {\bf Percent} \\ \hline
\textbf{\textit{ Disease NER}}  & 8270  & 819  & 9.90 \\ \hline
\textbf{\textit{Drug NER} }		& 9447  & 1309 & 13.85\\ \hline
\textbf{\textit{Clinical NER} }	& 13000 & 2617 & 20.13\\ \hline
\end{tabular}
}
\caption{Statistics of number of words not found in word embedding file. Here \textit{OoV} indicates number of words not found in pre-trained word embedding and \textit{percent} indicates its percentage over all vocabulary.}
\label{tab:oov}
\end{table}
 
In contrast to PoS tag embedding, we observe that use of pre-trained word embedding is the one of the important feature type in our model for each task. Pre-trained word embedding helps model to get better representation for rare words in training dataset.
 
\subsection{Effects of CRF and BLSTM}
We also analyze the unified framework to gain insight on the effect of using different loss function in the output layer (CRF vs. WLL) as well as effect of using bi-directional or uni-directional (forward) LSTM. For this analysis, we modify our framework and named model variants as follows: bi-directional LSTM with WLL output layer is called \textit{BLSTM+WLL} and uni-directional or regular LSTM with WLL layer is called \textit{LSTM+WLL}. In other words {\it BLSTM+WLL} model uses all the features of the proposed framework except it uses WLL in place of CRF. Similarly {\it LSTM+WLL} also uses all features along with forward LSTM instead of bidirectional LSTM and WLL in place of CRF. Results are presented in table \ref{tab:effect_model}. A relative decrement of $7.5\%$, $3.4\%$ and $5.5\%$ in obtained F Score on {\it Disease NER}, {\it Drug NER} and {\it Clinical NER} respectively by {\it BLSTM+WLL} compared to the proposed model demonstrate the importance of using CRF layer. This suggests that identifying tag independently is not favored by the model and it is better to utilize the implicit tag dependency. Further observation of average token length of a entity in three tasks indicates plausible reason for difference in performance in the three tasks. Average token length are 1.2 for drug entities, 2.1 for clinical and 2.2 for disease named entities. The longer the average length of entities, better the performance of model utilizing tag dependency. Similarly relative improvements of $12.89\%$, $4.86\%$ and $20.83\%$ in F1 score on {\it Disease NER}, {\it Drug NER} and {\it Clinical NER} tasks respectively are observed when compared with {\it LSTM + WLL}. This clearly indicates that the use of bi-directional LSTM is always advantageous.

% Since most of the drug names are short (in our dataset average 1.2 tokens for drug entity, 2.2 tokens disease entity, 2.1 tokens clinical entity) and dependency among tags happen more in multi token entities, this improvement are little less in {\it Drug NER} compare to other two tasks. 

\begin{table}[tbp]
\centering
\scalebox{0.8}{
\begin{tabular}{|l|c|c|c|}
\hline
{\textbf{Model}} & {\textbf{Disease NER}} & {\textbf{Drug NER}} & {\textbf{Clinical NER}} \\ 
\hline 
\hline
\textbf{\textit{CWBLSTM}} & (84.42), (82.31), (83.35) & (72.57), (74.60), (73.57) & (84.17), (83.20), (83.68)  \\ \hline
\textbf{\textit{BLSTM+WLL}} & (76.04), (78.25), (77.13) & (71.81), (70.34), (71.07) & (77.35), (80.91), (79.09) \\ \hline
\textbf{\textit{LSTM+WLL}}  & (64.72), (77.32), (70.46) & (68.41), (69.02), (68.71) & (58.32), (68.11), (62.83)\\ 
\hline
\end{tabular}
}
\caption{Effect of using CRF and WLL in output layer on performance of the proposed model on different datasets. In every block (X), (Y), (Z) implies precision, recall and F1 score respectively.}
\label{tab:effect_model}
\end{table}

\begin{table*}[h]
\begin{minipage}{\textwidth}
\centering
\scalebox{0.8}{
\begin{tabular} 
{|p{0.25\linewidth}|p{0.45\linewidth}|p{0.45\linewidth}|} \hline
\textbf{Word} & \textbf{Char BLSTM} & \textbf{GloVe} \\ \hline 
%{\bf aluminum} &  acetophenazine, diltiazem decamethonium, Minimum, amphetamine & copper, metal, cobalt, silver, metals  \\ \hline

%{\bf hydrochlorothiazide} & oxymetholone, chlorothiazide, Hydroxychloroquine, levothyroxine, chlortetracycline & amlodipine, lisinopril, atenolol, enalapril, valsartan \\ \hline

{\bf 2C19} &  2C9, 2C8/9, 29, 28.9, 2.9z & NA \\ \hline

{\bf synergistic} & septic, symptomatic, synaptic, serotonergic, synthetic & synergism, synergy, antagonistic, dose-dependent, exerts\\ \hline

{\bf dysfunction} & dysregulation, desensitization, dissolution, addition, admistration & impairment, impaired, disturbances, deterioration, insufficiency \\ \hline 

{\bf false-positive} & false-negative, facultative, five, folate, facilitate & false, falsely, erroneous, detecting, unreliable \\ \hline

{\bf micrograms/mL}  & microg/mL micromol/L micrograms/ml mg/mL mimicked &  NA \\ \hline
\end{tabular}
}
\caption{Word and its 5 nearest neighbors (from left to right increasing order of euclidean distance) learned by character level word embedding of our model on {\it Drug NER} corpus. NA implies word is not present in list}
\label{tab:emb_neg}
\end{minipage}
\end{table*}

\subsection{Analysis of Learned Word Embeddings}
Next we analyze characteristics of learned word embeddings after training of the proposed model. As mentioned earlier, we are learning two different representations of each word, one through its characters and other through distributional contexts. Our expectation is that the word embedding obtained through character embeddings will focus on morphological aspects whereas distributional word embedding on semantic and syntactic contexts.

We obtain character based word embedding for each word of the \textit{Drug NER} dataset after training. We picked $5$ words from test set of vocabulary list and observe its $5$ nearest neighbors in vocabulary list of training set. The nearest neighbors are selected using both word-embeddings and results are shown in Table~\ref{tab:emb_neg}. We can observe that the character based word embedding primarily focus on morphologically similar words, whereas distributional word embeddings preserve semantic properties. This clearly suggests that it is important to use the complementary nature of the two embeddings.

\section{Conclusion}
\label{conclusion}
In this research we present a unified model for drug, disease and clinical entity recognition tasks. Our model, called CWBLSTM, uses BLSTMs in hierarchy to learn better feature representation and CRF to infer correct labels for each word in the sentence at once. We believe, to the best of our knowledge, this is the first work using character based embeddings in drug and clinical entity recognition tasks. CWBLSTM outperforms task specific as well as task independent baselines in all three tasks. Through various analyses we demonstrated the importance of each feature type used by CWBLSTM. Our analyses suggest that pre-trained word embeddings and character based word embedding play complementary roles and along with incorporation of tag dependency are important ingredients for improving the performance of NER tasks in biomedical and clinical domains.
\section*{References}
\bibliography{acl2016}
 
\end{document}